\documentclass[11pt,a4paper]{article}
\usepackage{amssymb}
\usepackage{graphicx}
\usepackage{latexsym}
\usepackage{marvosym}
\usepackage{algorithmic}
\usepackage[ruled,vlined,linesnumbered]{algorithm2e}
\usepackage{amsfonts}
\usepackage{booktabs}
\usepackage{siunitx}
\usepackage{multirow}
\usepackage{rotating}
\usepackage{hhline}
\usepackage{dsfont}
\usepackage{amsmath}
\usepackage{enumerate}

\newcommand{\cutoff}{{\tt cutoff}}

\SetKw{myproc}{procedure}
\SetKw{myfunc}{function}
\SetKw{Inp}{in}
\SetKw{Outp}{out}
\SetKw{InOutp}{in out}
\SetKw{F}{false}
\SetKw{T}{true}
\SetKw{lAnd}{and}
\SetKw{Comment}{$\triangleright$\ }{}

\usepackage{etoolbox}
\newbool{seeAll}
\usepackage[usenames,dvipsnames]{color}
\usepackage[normalem]{ulem}
\usepackage{authblk}
\usepackage{xcolor}
\usepackage{soul}
\usepackage[utf8]{inputenc}
\usepackage[small]{caption}

\newcommand{\method}{{\sc Learn\&Optimize}\xspace}
\newcommand{\lang}{$\mathcal{L}$}
\newcommand{\basis}{$\mathcal{B}$}
\newcommand{\quacq}{\textsc{QuAcq}\xspace}

\makeatletter
\newcommand{\verbatimfont}[1]{\renewcommand{\verbatim@font}{\ttfamily#1}}
\usepackage{lineno,hyperref}

\title{Solve Optimization Problems with Unknown Constraint Networks}

\author[1]{Mohamed-Bachir Belaid}
\author[1]{Arnaud Gotlieb}
\author[2]{Nadjib Lazaar}
\affil[1]{Simula Research Laboratory, Oslo, Norway}
\affil[2]{LIRMM, University of Montpellier, CNRS, Montpellier, France}
\affil[ ]{\textit {\{bachir@simula.no,  arnaud@simula.no, lazaar@lirmm.fr\}}}
\date{}

\begin{document} 
\maketitle

\begin{abstract}
In most optimization problems,  users have a clear understanding of the function to optimize (e.g., minimize the makespan for scheduling problems). 
However, the constraints may be difficult to state and their modelling often requires expertise in Constraint Programming.
Active constraint acquisition has been successfully used to support non-experienced users in learning constraint networks through the generation of a sequence of queries. 
In this paper, we propose \method{},  a method to solve optimization problems with known objective function and unknown constraint network. 
It uses an active constraint acquisition algorithm
 which learns the unknown constraints and computes boundaries for the optimal solution during the learning process.
As a result, our method allows users to solve optimization problems without learning the overall constraint network.

\end{abstract}

\section{Introduction}
Constraint Programming (CP) is a framework to model and solve satisfaction and optimization problems.
Modeling problems into constraints is a crucial step in CP that usually requires expertise rarely available. 
In order to ease the creation of CP models, 
Bessiere et.al introduced constraint acquisition in  \cite{bessiere2017constraint} 
with the aim to learn CP models based on a set of assignment instances (passive learning) 
or specific queries that help to classify assignments (active learning). 

More precisely, sevral techniques and tools are available for constraint acquisition.
{\sc Conacq} \cite{bessiere2005sat} learns constraints that correctly classify a given set of positive (solutions) and negative (non-solutions) assignments in a passive way.
 {\sc Quacq} \cite{bessiere2013constraint} is an active learning algorithm that accepts or rejects constraints through interacting with the user with queries formed of partial assignments.
 {\sc Quacq} was further extended to learn all possible constraints from a negative query in \cite{arcangioli2016multiple}, to speed-up queries generation in \cite{addi2018time} 
and to exploit the structure of learned problems in \cite{tsouros2019structure}.
Other approaches for learning constraint models have also been developed, in particular to acquire global constraints such as in ModelSeeker \cite{beldiceanu2012model}.

Learning of optimization models is also a topic that received recently a lot of attention in the literature.
Elmachtoub et.al propose a general framework called \emph{Smart Predict then Optimize} (SPO) where parameters to optimization problems are predicted using advanced machine learning methods \cite{DBLP:journals/corr/abs-1710-08005}.
Similarly, Mandi et al.  proposed {\sc Predict\&Optimize}, a method that solves optimization problems where weights for the objective function are unknown in \cite{mandi2020smart}.

Since problems are usually modeled up to the point where they are solved, Bessiere et.al introduced in \cite{bessiere2014solve} a method where the acquisition process stops 
once a complete positive assignment is found. 
In this paper we extend the method of \cite{bessiere2014solve} to solve optimization problems
where the function to optimize is known and the constraints have to be learned.
Our idea is inspired from \cite{klindworth2012learning} where a precedence graph for the assembly line balancing problem is learned from past feasible (i.e., only positive) production sequences.
During the learning process the method attempts to find the optimal solution using the upper-bound (solution to the maximum graph) and the lower-bound (solution to the minimum graph).\footnote{See Figure 3 in \cite{klindworth2012learning} for more details.} 
Results in  \cite{klindworth2012learning} (and latter in \cite{otto2014multiple}) confirm that,  in many cases, the bounds are equal or very close to each other.
This method is limited to solving a specific problem (i.e.,  the assembly line balancing problem) and the learning part 
is limited to using feasible production sequences, to remove inconsistent precedence relations, and to direct interviews to confirm mandatory relations.

In this paper, we propose, \method{}, a general solution that uses
an active constraint acquisition method (e.g., {\sc Quacq}) for the constraint learning part.
During the acquisition process,   \method{} computes upper and lower bounds for the optimal value.
If these two bounds are equal, then optimality is proven. If these boundaries are close, we can say that a good-enough quality solution is found.
This work falls under the interactive optimization framework, where users are allowed to interact with the system
to improve solutions and find optimal ones quickly (please refer to \cite{meignan2015review} for a general overview).

\section{Learn\&Optimize}
In this section, we introduce \method, a method to solve optimization problems with unknown constraints.

\subsection{Notations}
The learner and the user share a common knowledge to communicate altogether, which is materialized by the vocabulary $(X,D,f)$ 
where $X$ is a set of $n$ variables, $D$ a specified finite domain, and $f$ an objective function on $X$.\footnote{Throughout the paper,  we assume that it is a minimization problem.}
In addition to the vocabulary, the learner owns the language  \lang{}, from which it can build the basis,  \basis{}, a set of constraints 
on specified sets of variables from $X$.
An assignment $e \in D^X$, 
is \emph{positive} (or feasible) if it satisfies the target network $T$, and it is \emph {negative} otherwise. 
We denote by $e^*$ an optimal solution to  "$f$  w.r.t. $T$".
Here, we assume that the user is always right and that $T \subseteq \mathcal{B}$. 
 
\subsection{\method{} Description}

\method{} is presented in Algorithm \ref{alg:one}.
It takes as input the vocabulary $(X,D,f)$,  the language \lang , a time bound \cutoff{}, a feasible solution $e_p$,\footnote{Handcrafted,  simple  to generate, but not necessary a solution of good quality.} and a precision value $\epsilon$.
\method{} returns the bounds $(lb, ub)$ and two assignments characterizing the bounds $(e_l,e_u)$.
First, the basis is created in line~\ref{lineAlg:initB} on $(X, \mathcal{L})$ w.r.t. the given feasible $e_p$.
Here, we keep in $\mathcal{B}$ only constraints accepting $e_p$.
That is, having a feasible solution $e_p$ is sufficient to guarantee that constraints in \basis{} accept at least one solution, while  $T\subseteq \mathcal{B}$ holds.
Then, the learned network $L$ is initialized to the empty set and the flag $optimal$ to $false$ (lines~\ref{lineAlg:initCL}-\ref{lineAlg:initF}).
\method{} loops until an optimal solution w.r.t. a given precision $\epsilon$ is found $(up-lb\leq\epsilon)$, the optimal solution to $(L)$ given $f$ is classified positive, or the learning and optimization time reaches the \cutoff{} time bound (line~\ref{lineAlg:cutoff}).
At each iteration, we generate an example to submit to the user (line~\ref{lineAlg:generate}).
If the submitted example $e$ is feasible (i.e., $e\models T$), we reduce $\mathcal{B}$ by removing all constraints not accepting $e$ (line~\ref{lineAlg:reduce}).
If the user says $no$,  a learning process based on \quacq{} is called, where a constraint is learned from a negative example (line~\ref{lineAlg:learn}).
Doing so, \method{} ensures to change the state of the basis $\mathcal{B}$ and/or the learned network $L$ 
and to have new bounds $(lb,ub)$ at each iteration.
Note that the lower and the upper bounds correspond to objective values of optimal solutions $e_l$ and $e_u$ of, respectively,  $(L)$ and $(L\cup \mathcal{B})$ given $f$ (lines \ref{lineAlg:e}-\ref{lineAlg:ublb}).
If the process reaches a precision of $\epsilon$ or $e_l$ is classified as positive (line~\ref{lineAlg:if}),  \method{} returns the optimal solution ($optimal = true$). Otherwise, \method{}  reaches the time bound value of \cutoff{}
 and returns a near-optimal solution.
%


\begin{algorithm}[h]
	\caption{ \method: Solve an optimization problem with known objective function, $f$, and unknown constraint network, $T$.}\label{alg:one} 
	\SetKw{Input}{In}
	\SetKw{InOut}{InOut}	
	\SetKw{In}{In}
	\SetKw{Out}{Out}
	\SetKw{InOut}{InOut}	
	\SetKw{Output}{Output}
	\SetKw{return}{return}
	\SetKwFor{Do}{do}{}{}
	\DontPrintSemicolon
	\BlankLine
	\In:  $(X,D,f)$; \lang ; \cutoff ; $e_p$; $\epsilon$;
	
	\Out: $((e_l,lb),(e_u,ub))$;
	\BlankLine
	
	\Begin{
		$\mathcal{B} \gets {\tt CreateBasis}(X,\mathcal{L},e_p);$\\ \label{lineAlg:initB}
		$L \gets \emptyset$;\\ \label{lineAlg:initCL}
		$optimal\gets false$;\\ \label{lineAlg:initF}
		\Do{in $t<\cutoff$}{ \label{lineAlg:cutoff}
		\While{$\neg optimal$ }	{ \label{lineAlg:loop}
			$e\gets {\tt queryGenerator}(X,D,L,\mathcal{B})$;\ \label{lineAlg:generate}
			
			\lIf{$ask(e)=yes$}{
			${\tt reduce}(\mathcal{B},e);$}\label{lineAlg:reduce}
		\lElse{
		${\tt learn}(L,\mathcal{B},e);$} \label{lineAlg:learn}
		$e_l\gets OptSol(L, f)$;
				$e_u\gets OptSol(L\cup\mathcal{B}, f)$;\\ \label{lineAlg:e}		
		$lb\gets f(e_l)$;
		$ub\gets f(e_u)$;\\\label{lineAlg:ublb}
		\If{$((ub-lb)\leq\epsilon)$ \textbf{or} $(ask(e_l)=yes)$}{
		\label{lineAlg:if} $optimal\gets true$;\\\label{lineAlg:fupdate}
		} 
		}
	}
			\return{$((e_l,lb),(e_u,ub))$;}\label{lineAlg:return}			
		}	
		
\end{algorithm}

\section{Conclusion}
In this paper, we have presented, \method, a method to solve optimization problems in the context where the optimization function is known and the constraint network is not. 
Some evidence from the literature show that optimal solutions can be found without necessary fully modelling the problem \cite{klindworth2012learning}.
In our future work, we will further investigate the application of \method{} to some well-chosen real world problems.

\bibliographystyle{abbrv} 

\bibliography{paper}

\end{document}